\PassOptionsToPackage{table}{xcolor}
\documentclass{fairmeta}
\usepackage[utf8]{inputenc}

\usepackage{amsmath,amsfonts,bm}

\def\eqref#1{equation~\ref{#1}}

\def\1{\bm{1}}

\DeclareMathAlphabet{\mathsfit}{\encodingdefault}{\sfdefault}{m}{sl}
\SetMathAlphabet{\mathsfit}{bold}{\encodingdefault}{\sfdefault}{bx}{n}

\usepackage{amssymb}
\usepackage{array,tabularx}
\usepackage{url}
\definecolor{citationblue}{HTML}{4F7294}
\hypersetup{colorlinks=true,citecolor=citationblue,linkcolor=black,urlcolor=black,filecolor=black,
  pdftitle={Frontier-OPD: Discovering the Edge of Agent Capability and Teaching Agents to Go Beyond},
  pdfauthor={Zheng Jiang, Yiming Chen, Houde Qian, Ling Li, Chaoyang Li, Lifeng Sun},pdfsubject={On-policy distillation for multi-turn agents}}

\definecolor{FrontierCream}{HTML}{F7EDE1}
\definecolor{FrontierLavender}{HTML}{EFE9F7}
\definecolor{FrontierBlue}{HTML}{EBF3FC}
\definecolor{FrontierMint}{HTML}{DFEFE5}

\title{Frontier-OPD: Discovering the Edge of Agent Capability and Teaching Agents to Go Beyond}

\renewcommand{\authorformat}[2][]{{\sffamily\large\bfseries #2\ifstrempty{#1}{}{\textsuperscript{\ensuremath{#1}}}}}
\renewcommand{\affiliationformat}[2][]{{\sffamily\normalsize\ifstrempty{#1}{}{\textsuperscript{\ensuremath{#1}}}#2}}
\renewcommand{\contributionformat}[2][]{{\sffamily\normalsize\ifstrempty{#1}{}{\textsuperscript{\ensuremath{#1}}}#2}}
\author{Zheng Jiang}
\author{Yiming Chen}
\author{Houde Qian}
\author{Ling Li}
\author{Chaoyang Li}
\author[\dagger]{Lifeng Sun}
\affiliation{Tsinghua University}
\contribution[\dagger]{Corresponding author}
\metadata[Code]{\url{https://github.com/jiangz20/FrontierOPD}}
\metadata[Contact]{\email{jz24@mails.tsinghua.edu.cn}, \email{sunlf@tsinghua.edu.cn}}

\abstract{
On-policy distillation aligns teacher feedback with a student's current
behavior, yet acquiring useful supervision for multi-turn agents requires
identifying both where guidance is needed and whether the student can
sustain the interaction afterward. We introduce \textbf{Frontier-OPD},
which addresses this problem through the student's \emph{learning frontier}:
decision situations that remain challenging but admit useful autonomous
continuation after brief teacher guidance. To estimate this frontier,
Frontier-OPD organizes on-policy experience into a shared-state trajectory
graph, where trajectory-balanced aggregation of teacher--student
disagreement identifies distinct intervention candidates while preserving
the histories needed for replay. Short teacher bridges are then assessed
by comparing agreement over the bridge and ensuing student continuation
with an equally long segment of the original trajectory, providing a
behavioral proxy for guidance that the current student can build on.
Accepted teacher responses augment ordinary distillation data, and fresh
rollouts renew the frontier estimate as the policy changes. Experiments
on ALFWorld, WebShop, and ScienceWorld report higher mean autonomous
performance than the compared distillation baselines across the evaluated
student scales and teacher configurations, with ablations indicating
complementary effects of candidate selection and guidance filtering. In one
ALFWorld timing setting, the additional collection cost is offset by a
shorter time to a target success rate.
}

\begin{document}
\maketitle
\pagestyle{plain}
\thispagestyle{empty}

\section{Introduction}
\label{sec:introduction}

Multi-turn agents interact with their environments through sequences of
decisions whose consequences determine the observations and choices that
follow~\citep{yao2023react}, making autonomous task completion dependent
on coordinated behavior across the entire interaction. Distilling this
ability therefore requires supervision that remains relevant to the
histories induced by the student's own policy, a concern closely related
to distribution shift in interactive imitation learning~\citep{ross2011dagger}.
On-policy distillation (OPD) addresses this concern by querying a teacher
on student-generated responses~\citep{agarwal2024opd}; however, aligning
feedback with current behavior leaves open the question of which additional
teacher guidance is most appropriate for the student to learn from next.

Teacher--student disagreement alone provides an incomplete basis for
answering this question, because the need to correct a decision and the
student's ability to use that correction are distinct considerations.
Where the student already behaves reliably, additional guidance may offer
limited benefit; where its behavior differs substantially from the
teacher's, an intervention may still leave it facing subsequent decisions
that exceed its current capabilities. Consider a student that can complete
a placement task once shown a particular object manipulation: a short
demonstration at that decision may provide usable guidance, whereas an
intervention leading to an unfamiliar procedure may leave the student
unable to sustain the interaction. Effective supervision acquisition thus
requires identifying learning needs that can be addressed through guidance
compatible with the student's subsequent autonomous behavior.

We characterize these opportunities as the student's \emph{learning
frontier}: decision situations that it has not reliably mastered, but from
which brief teacher guidance can enable useful autonomous continuation. This frontier is relative to the
current policy and concerns the suitability of a learning opportunity,
rather than simply whether an environment state has been visited.
Because a familiar state may still require an unfamiliar decision, while
a newly encountered state need not represent an acquired skill, experience
coverage alone cannot determine which guidance the student should receive.
As learning changes both its behavior and its ability to continue from
teacher interventions, the frontier must be reassessed accordingly.

\begin{figure}[!t]
  \centering
  \includegraphics[width=\textwidth]{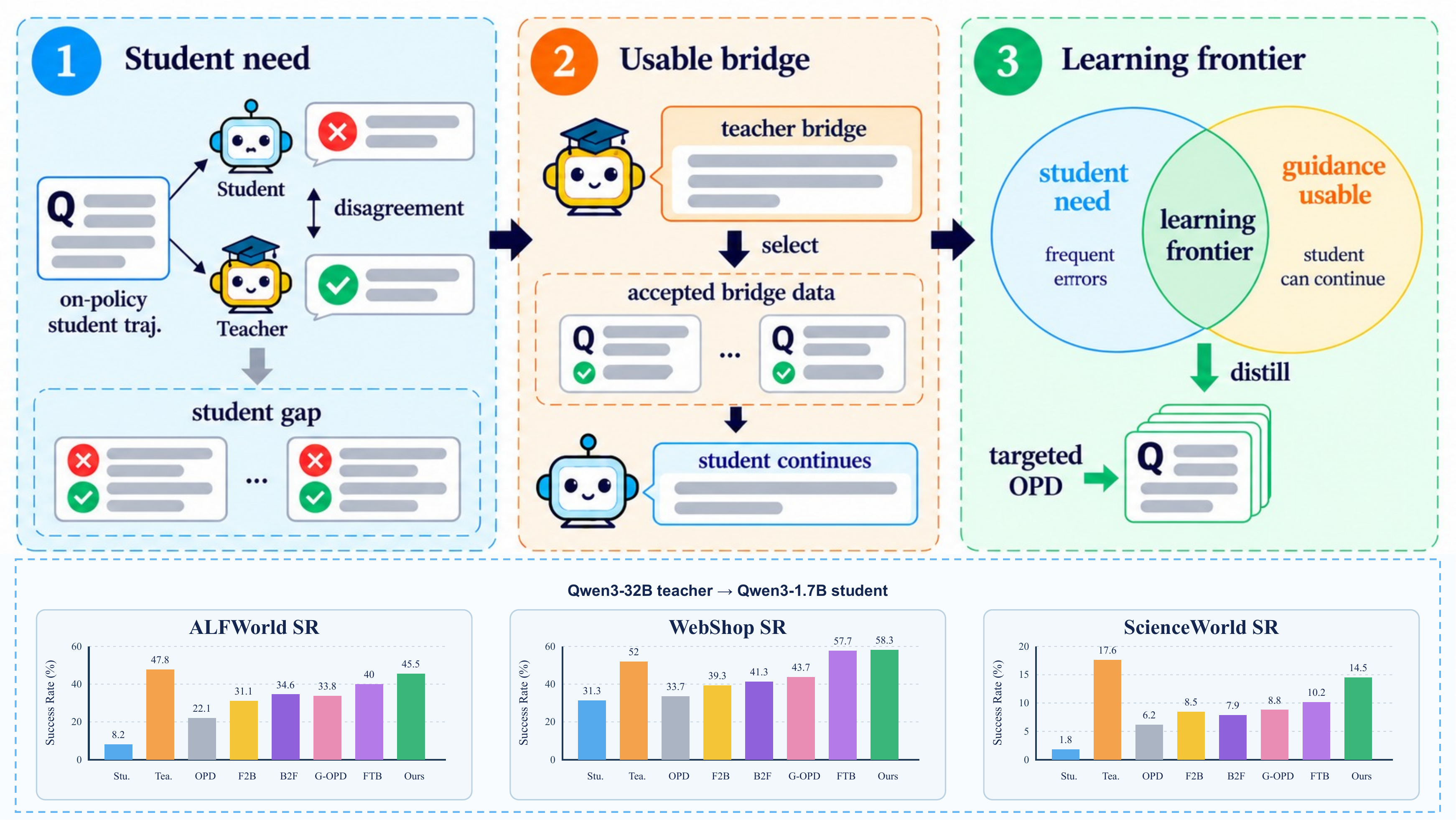}
  \caption{Motivation for Frontier-OPD. Repeated student--teacher disagreement
  is treated as evidence of a potential learning need; a teacher bridge may be
  useful when it enables the student to continue; their overlap is
  operationalized as the learning frontier used for targeted distillation.
  The shared results strip reports success rate
  (
  highlighting the student, teacher, and Frontier-OPD outcomes.}
  \label{fig:learning_frontier}
\end{figure}

Estimating this frontier entails two coupled challenges, illustrated in
Figure~\ref{fig:learning_frontier}.
\emph{Identifying learning needs} requires relating individual responses
to the decision situations in which they arise: an isolated large
teacher--student gap may not represent a persistent weakness, and repeated
visits within one attempt can disproportionately influence the evidence.
The representation must consequently consolidate behavior across shared
situations while preserving the histories that condition individual
actions. \emph{Assessing the usability of guidance}, in turn, requires
evidence from the student's behavior after control is returned, since the
quality of the teacher's action does not by itself establish whether the
student can continue productively from the resulting history.

Recent agentic distillation methods address complementary aspects of
supervision acquisition: TCOD varies student execution depth through
temporal curricula~\citep{wang2026tcod}, Guided-OPD schedules teacher
participation across turns~\citep{li2026guidedopd}, and FutureBridge-OPD
validates localized guidance through subsequent student
continuations~\citep{chen2026futurebridge}. Building on continuation-based
validation, we investigate how decision situations should be selected for
intervention and how that selection can be coupled to an assessment of
the resulting guidance. The learning-frontier perspective connects these
two decisions through the evolving needs and capabilities of the student.

We introduce \textbf{Frontier-OPD}, which operationalizes and tracks a proxy
for the learning frontier by combining shared-state evidence aggregation with
history-specific intervention and validation. Student trajectories are
organized into a graph whose shared nodes align observed decision
situations, allowing trajectory-balanced aggregation of disagreement to
rank distinct \emph{frontier candidates} without allocating multiple
candidate slots to repeated visits at the same node. Each candidate is
replayed from a concrete history, after which a short teacher bridge is
assessed by comparing agreement over the bridge and ensuing student
continuation with an equally long original segment. These behavioral
proxies guide the acquisition of teacher responses for distillation,
while fresh student rollouts renew the candidate pool and scores as the
policy changes.

Our contributions are threefold:
\begin{itemize}
\item We formulate a learning-frontier perspective on agentic OPD that
connects evidence of unmet learning needs with the student's ability to
continue from brief teacher guidance.
\item We develop Frontier-OPD, which combines trajectory-balanced selection
of shared-state candidates with history-specific replay and matched-window
validation to adapt supervision acquisition to the current student.
\item We evaluate the approach on ALFWorld, WebShop, and ScienceWorld across
student scales and teacher configurations, reporting higher mean autonomous
performance in the evaluated comparisons and examining the selection and
validation procedure through ablations, budget sensitivity, and training-time
comparisons.
\end{itemize}

\section{Related Work}
\label{sec:related_work}

\paragraph{On-Policy Distillation for Multi-Turn Agents.}
Knowledge distillation transfers teacher predictions to a student
\citep{hinton2015distilling}, while on-policy approaches align supervision
with student-generated responses~\citep{agarwal2024opd} and support
reverse-KL optimization for generative models~\citep{gu2024minillm}.
In multi-turn environments, the student's actions also determine the
histories on which later feedback is obtained, making experience
construction an important part of the training procedure. Drawing on
curriculum principles that adapt learning opportunities to progress
\citep{graves2017curriculum,florensa2017reverse}, TCOD regulates student
execution depth~\citep{wang2026tcod}, whereas Guided-OPD schedules teacher
participation within rollouts~\citep{li2026guidedopd}. Frontier-OPD retains
student-driven collection while organizing observed decision situations
as a shared reference for acquiring additional supervision.

\paragraph{Selective Teacher Guidance and Validation.}
Beyond constructing training trajectories, interactive learning must
determine where expert guidance is warranted and how its usefulness
should be assessed. DAgger collects expert labels under learner-induced
state distributions~\citep{ross2011dagger}, while ThriftyDAgger uses
novelty and risk to allocate interventions under a budget
\citep{hoque2022thriftydagger}; in agentic distillation, FutureBridge-OPD
assesses short teacher bridges through subsequent student continuations
\citep{chen2026futurebridge}. These approaches motivate complementary
considerations of intervention need and downstream behavior. Our
learning-frontier perspective connects them through trajectory-balanced
aggregation of disagreement at shared states and matched-window
validation of history-specific guidance, adapting the supervision selected
for training to both the student's observed behavior and its continuation
after teacher intervention.

\section{Method}
\label{sec:method}

\begin{figure}[htbp]
  \centering
  \includegraphics[width=\textwidth]{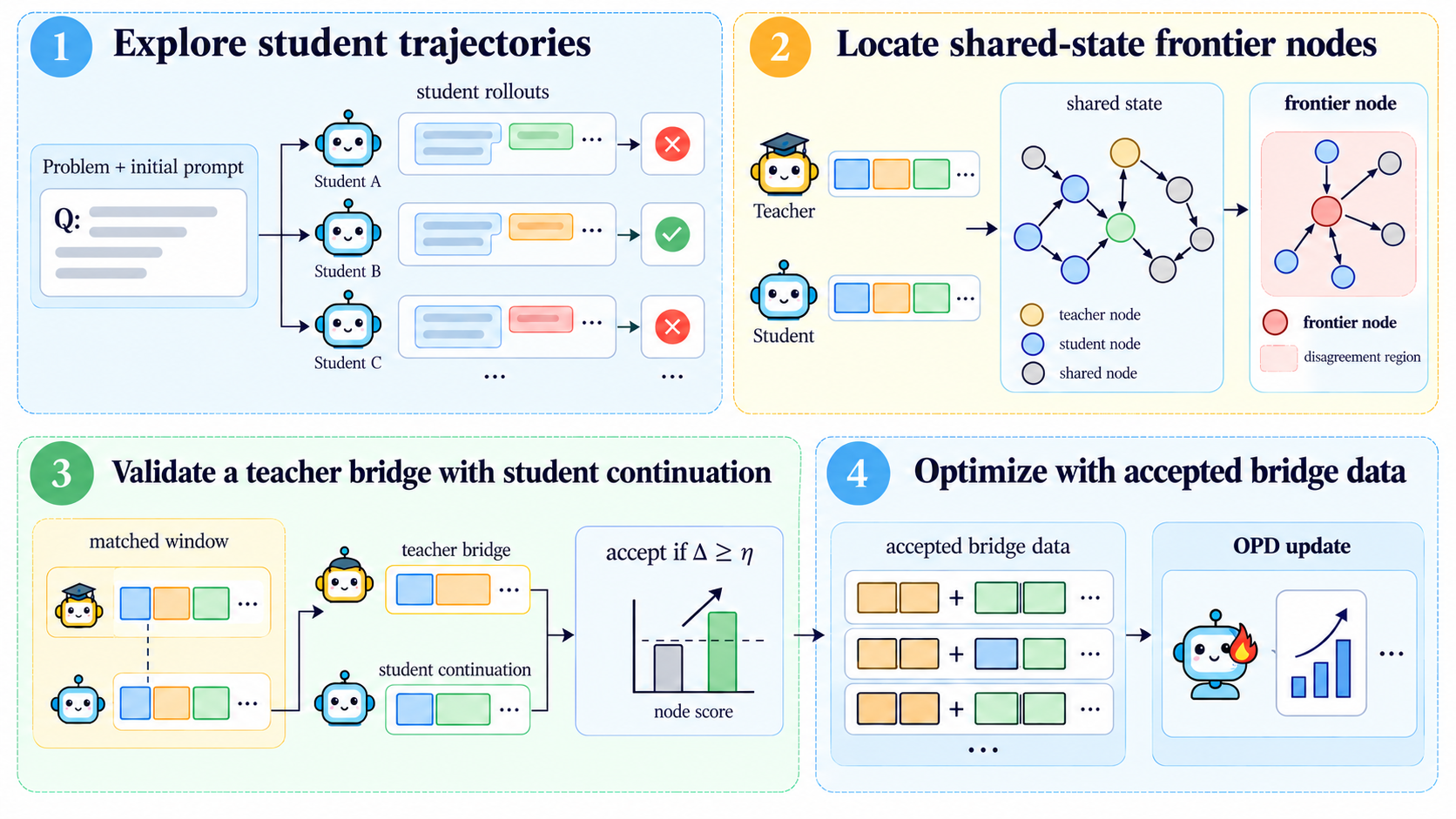}
  \caption{Overview of Frontier-OPD. Trajectory-balanced disagreement
  aggregation ranks shared-state frontier candidates, each of which supplies
  a concrete history for teacher intervention. Matched-window validation
  assesses the bridge and subsequent student behavior before accepted
  teacher responses enter distillation, after which fresh rollouts renew
  the frontier estimate.}
  \label{fig:frontier_overview}
\end{figure}

Frontier-OPD estimates the learning frontier through two complementary
operations: aggregating teacher--student disagreement to identify candidate
learning needs, and validating short teacher interventions through the
student's subsequent behavior (Figure~\ref{fig:frontier_overview}). A
shared-state trajectory graph provides the reference for organizing
behavioral evidence, while the original decision histories are retained
so that each candidate can be replayed and evaluated in its own context.
This distinction allows supervision to be allocated at the level of a
shared decision situation without replacing the history on which the
student and teacher condition their responses.

Let $\pi_{\theta_k}$ and $\pi_{\mathrm T}$ denote the student at collection
iteration $k$ and the teacher, respectively. At turn $t$, the agent
conditions on interaction history $h_t$ and generates response $y_t$
specifying environment action $a_t$; shared state identities are used for
evidence aggregation and ranking, whereas the full histories determine
intervention replay and validation.

\subsection{Representing Learning Needs in Student Experience}
\label{sec:frontier_estimation}

\paragraph{Student experience.}
For task $x$, we collect $M$ trajectories
$\mathcal{R}_k(x)=\{\tau_k^i\}_{i=1}^{M}$ by sampling student responses
on-policy from prescribed starting histories. These histories consist of
the scheduled reference prefix when an expert-prefix curriculum is active
\citep{wang2026tcod}, or the task's initial state otherwise, with the same
initialization and history-construction rules maintained within each
controlled comparison.

\paragraph{Empirical trajectory graph.}
To align behavioral evidence across visits, we construct a task-specific
directed graph $\mathcal{G}_k(x)=(\mathcal{V}_k,\mathcal{E}_k)$ by mapping
each observation $o_t^i$ and its available-action information
$\mathcal{A}_t^i$ to a state key
$v_t^i=\phi(o_t^i,\mathcal{A}_t^i)$, where $\phi$ hashes their normalized
representations. The resulting vertices and action-labeled edges are
\begin{equation}
  \begin{aligned}
  \mathcal{V}_k &= \bigcup_{i,t}\{v_t^i\}, \\
  \mathcal{E}_k &= \bigcup_{i,t}\{(v_t^i,a_t^i,v_{t+1}^i)\}.
  \end{aligned}
  \label{eq:trajectory_graph}
\end{equation}
Shared keys merge repeated observations within and across trajectories,
while the recorded transitions preserve observed connectivity, yielding
an empirical representation of behavioral coverage under an approximate
observational identity. Since a shared node can be reached through
different histories, each decision occurrence $c=(i,t)$ retains its
original context, response, and replay prefix; the recorded sequences and
shared keys therefore provide an implicit graph without collapsing the
contexts used for generation. State identity determines the grouping used
in node-score aggregation and whether a teacher transition reaches a
previously observed state, whereas graph adjacency does not enter the
ranking rule.

\paragraph{Sampled signed disagreement.}
For response token $y_j$ under context $(h,y_{<j})$, define the recorded
student--teacher log-probability gap as
\begin{equation}
  \begin{aligned}
  g_k(h,y,j) ={}& \log\pi_{\theta_k}(y_j\mid h,y_{<j}) \\
               &-\log\pi_{\mathrm{T}}(y_j\mid h,y_{<j}).
  \end{aligned}
  \label{eq:token_gap}
\end{equation}
For occurrence $c=(i,t)$, we use sampled signed
disagreement~\citep{chen2026futurebridge} as the occurrence-level statistic:
\begin{equation}
  d_k(c)=\frac{1}{|y_t^i|}\sum_{j=1}^{|y_t^i|}
             g_k(h_t^i,y_t^i,j).
  \label{eq:signed_disagreement}
\end{equation}
Because higher values indicate that the student assigns a higher
log-probability to the sampled response than the teacher does, this signed,
response-level statistic provides a learning-need proxy that can be
aggregated across eligible visits before selecting intervention candidates.

\subsection{Selecting Learning-Frontier Candidates}
\label{sec:candidate_selection}

Only eligible decision occurrences contribute to candidate ranking.
Writing $\mathcal{Q}_k^i$ for their set in trajectory $i$, eligibility
requires valid actions through the candidate, a replayable prefix, a
state key, aligned nonempty student and teacher token scores, finite
$d_k(c)$, and at least $K+H$ recorded response turns starting at $c$.
The final requirement ensures that a reference segment is available for
a teacher bridge of at most $K$ turns followed by a student-validation
horizon of $H$ turns.

\paragraph{Trajectory-balanced node scores.}
For node $v$, define its eligible visits in trajectory $i$ and the set of
contributing trajectories as
\begin{equation}
  \begin{aligned}
  \mathcal{Q}_k^i(v)&=\{(i,t)\in\mathcal{Q}_k^i:v_t^i=v\}, \\
      \mathcal{I}_k(v)&=\{i:|\mathcal{Q}_k^i(v)|>0\}.
  \end{aligned}
  \label{eq:node_visits}
\end{equation}
We give each contributing trajectory equal total weight and distribute
that weight equally among its eligible visits to $v$:
\begin{equation}
  \begin{aligned}
  s_k(v)&=\sum_{i\in\mathcal{I}_k(v)}
           \sum_{c\in\mathcal{Q}_k^i(v)}w_k(c,v)\,d_k(c), \\
      w_k(c,v)&=\frac{1}{|\mathcal{I}_k(v)|\,|\mathcal{Q}_k^i(v)|}.
  \end{aligned}
  \label{eq:node_score}
\end{equation}
Equation~\eqref{eq:node_score} first averages disagreements within each
contributing trajectory and then averages those trajectory means, so
repeated visits can change a trajectory's estimate without increasing its
total weight at the node. Trajectories with no eligible visit contribute
no weight, and response length does not affect visit weights; this avoids
allowing a looping trajectory to dominate solely through its visit count,
while retaining nodes supported by only one eligible occurrence.

\paragraph{Distinct frontier candidates and replay representatives.}
Among eligible nodes
$\mathcal{U}_k=\{v\in\mathcal{V}_k:|\mathcal{I}_k(v)|>0\}$, we apply an
optional threshold $\delta$ to the aggregated score and select at most
$B$ distinct candidates:
\begin{equation}
  \mathcal{F}_k=\operatorname{TopB}_{s_k}
     \bigl(\{v\in\mathcal{U}_k:s_k(v)\geq\delta\}\bigr).
  \label{eq:candidate_selection}
\end{equation}
These \emph{frontier candidates} are prioritized by the learning-need
proxy before their guidance has been tested; to associate each selected
node with a concrete intervention context, we choose its
highest-disagreement eligible occurrence as the replay representative:
\begin{equation}
  \begin{aligned}
  c_k^*(v)&=\arg\max_{c\in\bigcup_i\mathcal{Q}_k^i(v)}d_k(c), \\
      \mathcal{C}_k&=\{c_k^*(v):v\in\mathcal{F}_k\}.
  \end{aligned}
  \label{eq:node_representative}
\end{equation}
Occurrence ties follow collection order (trajectory, then turn), and
node-score ties follow the first eligible visit. Because the budget applies
to distinct nodes rather than trajectories, several representatives may
come from the same trajectory, but repeated visits to one node cannot
consume multiple candidate slots.

For each $c\in\mathcal{C}_k$, the environment is reset and the original
prefix---including the expert prefix when present and the preceding
student actions---is replayed before the teacher generates a short bridge
from the saved decision context. Teacher control ends after at most $K$
turns, at the first state key outside $\mathcal{V}_k$, or upon environment
termination, giving a realized length $L_c\leq K$. Validation requires
a nonterminal handoff to the student; with the default $K=1$, both
graph-internal and graph-external handoffs are eligible, so graph exit
is not an acceptance requirement.

\subsection{Continuation-Validated Teacher Bridges}
\label{sec:continuation_validation}

Whereas candidate ranking identifies decision situations at which to test
guidance, validation assesses the resulting intervention after the student
resumes control. Following the continuation-based rationale of
\citet{chen2026futurebridge}, we generate $H$ student response turns after
the teacher bridge and compare the joint window against a matched segment
of the original trajectory. For this assessment, we use the negative
absolute token log-probability gap,
\begin{equation}
  a_k(h,y)=-\frac{1}{|y|}\sum_{j=1}^{|y|}|g_k(h,y,j)|.
  \label{eq:teacher_agreement}
\end{equation}
so larger values indicate closer agreement on the evaluated response,
unlike the signed statistic used for candidate localization. Window-level
agreement $A_k(W)$ is the arithmetic mean of $a_k(h,y)$ over the response
turns in $W$, with both models scoring each response---including
teacher-generated bridge responses---under its own context. Normalizing
within responses and then averaging uniformly across turns prevents
response length from determining its contribution to the window score.

The new window $W_c^{\mathrm{new}}$ contains the realized teacher bridge
\emph{and} its student continuation, while $W_c^{\mathrm{old}}$ comprises
the original $L_c+H$ response turns beginning at the same occurrence.
Reusing the recorded reference scores yields a comparison that preserves
the task and intervention prefix while matching the number of response
turns, with agreement gain defined as
\begin{equation}
  \Delta_k(c)=A_k(W_c^{\mathrm{new}})-A_k(W_c^{\mathrm{old}}).
  \label{eq:relative_gain}
\end{equation}
Let $V_k(c)$ indicate successful prefix replay, a nonterminal teacher handoff,
a complete valid $H$-turn student continuation, and an equally long reference
window with defined agreement scores. The accepted candidate set is
\begin{equation}
  \mathcal{B}_k=\{c\in\mathcal{C}_k:V_k(c)=1,\ \Delta_k(c)\geq\eta\}.
  \label{eq:accepted_bridges}
\end{equation}
With the default $\eta=0$, the filter retains agreement-preserving or
agreement-improving windows and contributes every accepted teacher bridge
to the supervision pool, while the newly generated student continuations
serve only as validation evidence. This criterion refines the candidate
set using a proxy for usable guidance; because it evaluates the joint
window, acceptance does not establish an improvement in the continuation
alone, task success, or learning after an update.

\subsection{Distillation and Learning-Frontier Renewal}
\label{sec:frontier_expansion}

Let $\mathcal{D}_k^{\mathrm{stu}}$ contain ordinary student response
experiences and $\mathcal{D}_k^{\mathrm{br}}$ contain teacher response
experiences from accepted bridges, each retaining its model context,
response tokens, and recorded student and teacher token log-probabilities.
For an experience $e$ with response tokens $y_{e,1:L_e}$, the workflow
stores the behavior-model log probability $\ell^{\mathrm{stu}}_{e,j}$ and
the teacher log probability $\ell^{\mathrm{T}}_{e,j}$ for every response
token. For ordinary rollouts, the behavior model is the sampling student;
for an accepted bridge, the teacher samples the response and the student
is evaluated on those teacher-sampled tokens. The multi-turn OPD trainer
uses the token-level OPD advantage
\begin{equation}
  \widehat A_{e,j}=\ell^{\mathrm{T}}_{e,j}-\ell^{\mathrm{stu}}_{e,j}
  \label{eq:opd_token_advantage}
\end{equation}
with the configured coefficient $\kappa=1.0$ in the actor update. The
bridge responses therefore change which contexts and tokens enter the
on-policy distillation update, but do not introduce a separate bridge loss
or a reward-weighted objective; their stored reward is unused by
\texttt{multi\_turn\_opd}. The newly generated student continuations used
for validation are not added to the training pool.

Both sources enter the same distillation update:
\begin{equation}
  \begin{aligned}
  \mathcal{D}_k &= \mathcal{D}_k^{\mathrm{stu}}\uplus\mathcal{D}_k^{\mathrm{br}}, \\
  \theta_{k+1} &= \operatorname{OPDUpdate}(\theta_k;\mathcal{D}_k).
  \end{aligned}
  \label{eq:coupled_update}
\end{equation}
Here $\uplus$ concatenates experience multisets without collapsing
repeated occurrences, and $\operatorname{OPDUpdate}$ denotes the
multi-turn OPD actor update~\citep{agarwal2024opd} with the same optimizer
for both sources. We use four repeated rollout samples per task, AdamW with
learning rate $10^{-6}$, default $(\beta_1,\beta_2)=(0.9,0.999)$, weight
decay $10^{-2}$, and gradient clipping at $1.0$. Their relative
contributions depend on the accepted data and experience sampling.
Subsequent on-policy rollouts reconstruct the graph and renew
its scores and replay representatives, allowing the learning-frontier
estimate to adapt to changes in both the student's behavior and its
continuation from teacher guidance. The resulting policy is evaluated
autonomously from the task's initial state, without teacher intervention
or reference-prefix replay.

\begin{table}[htbp]
\centering

\small
\setlength{\tabcolsep}{4pt}
\renewcommand{\arraystretch}{1.12}
\begin{tabular}{@{}p{0.24\linewidth}*{5}{>{\centering\arraybackslash}p{\dimexpr0.152\linewidth-2\tabcolsep\relax}}@{}}
\toprule
& \textbf{ALFWorld} & \multicolumn{2}{c}{\textbf{WebShop}}
& \multicolumn{2}{c}{\textbf{ScienceWorld}} \\
\cmidrule(lr){2-2}\cmidrule(lr){3-4}\cmidrule(l){5-6}
\textbf{Method} & SR $\uparrow$ & Score $\uparrow$ & SR $\uparrow$
& Score $\uparrow$ & SR $\uparrow$ \\
\midrule
\rowcolor{FrontierCream}
\multicolumn{6}{c}{\rule{0pt}{2.5ex}\textbf{Qwen3-32B teacher $\boldsymbol{\rightarrow}$ Qwen3-1.7B student}} \\
\addlinespace[3pt]
Student (zero-shot) & 8.2 & 40.5 & 31.3 & 8.3 & 1.8 \\
Teacher (zero-shot) & 47.8 & 54.0 & 52.0 & 45.1 & 17.6 \\
OPD & $22.1 \pm 2.8$ & $35.2 \pm 2.0$ & $33.7 \pm 2.9$
& $27.5 \pm 1.5$ & $6.2 \pm 1.0$ \\
TCOD-F2B & $31.1 \pm 3.0$ & $41.3 \pm 1.9$ & $39.3 \pm 1.5$
& $29.8 \pm 0.8$ & $8.5 \pm 0.9$ \\
TCOD-B2F & $34.6 \pm 3.1$ & $43.6 \pm 3.2$ & $41.3 \pm 3.1$
& $28.4 \pm 1.2$ & $7.9 \pm 0.6$ \\
Guided-OPD & $33.8 \pm 5.1$ & $46.4 \pm 10.5$ & $43.7 \pm 9.5$
& $30.2 \pm 1.8$ & $8.8 \pm 1.1$ \\
FutureBridge-OPD & $40.0 \pm 3.1$ & $59.0 \pm 1.6$
& $57.7 \pm 1.5$ & $32.6 \pm 1.0$ & $10.2 \pm 0.7$ \\
\addlinespace[1pt]
\rowcolor{FrontierBlue}
\textbf{Frontier-OPD (ours)} & $\mathbf{45.5 \pm 2.0}$ & $\mathbf{59.6 \pm 1.2}$
& $\mathbf{58.3 \pm 1.5}$ & $\mathbf{40.0 \pm 2.4}$ & $\mathbf{14.5 \pm 0.5}$ \\
\midrule
\rowcolor{FrontierLavender}
\multicolumn{6}{c}{\rule{0pt}{2.5ex}\textbf{Qwen3-32B teacher $\boldsymbol{\rightarrow}$ Qwen3-4B student}} \\
\addlinespace[3pt]
Student (zero-shot) & 21.1 & 33.0 & 29.3 & 23.5 & 8.5 \\
Teacher (zero-shot) & 47.8 & 54.0 & 52.0 & 45.1 & 17.6 \\
OPD & $42.8 \pm 4.5$ & $43.8 \pm 1.6$ & $40.7 \pm 2.3$
& $29.8 \pm 0.3$ & $8.2 \pm 0.9$ \\
TCOD-F2B & $47.0 \pm 4.9$ & $49.6 \pm 7.2$ & $46.3 \pm 6.4$
& $33.5 \pm 0.6$ & $12.1 \pm 2.2$ \\
TCOD-B2F & $40.3 \pm 4.7$ & $50.8 \pm 1.4$ & $47.3 \pm 2.9$
& $32.6 \pm 1.9$ & $11.2 \pm 0.7$ \\
Guided-OPD & $47.5 \pm 8.7$ & $48.5 \pm 5.5$ & $47.7 \pm 5.7$
& $34.8 \pm 1.6$ & $12.8 \pm 1.8$ \\
FutureBridge-OPD & $48.5 \pm 2.0$ & $54.2 \pm 1.2$
& $51.7 \pm 1.5$ & $38.2 \pm 1.1$ & $13.5 \pm 0.8$ \\
\addlinespace[1pt]
\rowcolor{FrontierBlue}
\textbf{Frontier-OPD (ours)} & $\mathbf{49.5 \pm 1.9}$ & $\mathbf{54.8 \pm 1.0}$
& $\mathbf{52.3 \pm 1.2}$ & $\mathbf{41.8 \pm 1.5}$ & $\mathbf{16.2 \pm 0.6}$ \\
\bottomrule
\end{tabular}
\caption{Performance comparison with Qwen3-32B as teacher and Qwen3-1.7B
or Qwen3-4B as student. Trained-policy results are mean $\pm$ sample
standard deviation over three seeds. Boldface marks the highest mean among
trained students within each configuration; blue shading highlights Frontier-OPD.}
\label{tab:main_results}
\end{table}

\section{Experiments}
\label{sec:experiments}

We evaluate whether supervision selected through the learning-frontier
procedure improves autonomous task performance, how candidate selection
and guidance filtering contribute, and how the collection budgets affect
performance and training time.

\subsection{Experimental Setup}
\label{sec:experimental_setup}

\paragraph{Benchmarks and metrics.}
We evaluate Frontier-OPD on ALFWorld for household interaction
\citep{shridhar2020alfworld}, WebShop for grounded web interaction
\citep{yao2022webshop}, and ScienceWorld for interactive scientific tasks
\citep{wang2022scienceworld}, reporting success rate (SR) and, where
available, task score. All evaluations use held-out task instances on
which the student executes independently from the environment's initial
state, following the protocol in Appendix~\ref{sec:evaluation_protocol}.

\paragraph{Models and comparisons.}
The main comparison uses Qwen3-32B as teacher with Qwen3-1.7B and Qwen3-4B
students~\citep{yang2025qwen3}, supplemented by an RL-trained Qwen3-8B
teacher (Qwen3-8B-RL) with the same student scales to examine transfer across teacher
configurations (Section~\ref{sec:teacher_configuration}). We compare
against OPD~\citep{agarwal2024opd}, TCOD-F2B and TCOD-B2F
\citep{wang2026tcod}, Guided-OPD~\citep{li2026guidedopd}, and
FutureBridge-OPD~\citep{chen2026futurebridge}, with zero-shot student and
teacher policies providing reference performance. Within each controlled
comparison, we fix initialization, training tasks, reference-prefix
curriculum, prompts, decoding, and optimization settings; the reference
Frontier-OPD configuration uses $M=4$, $B=4$, $K=1$, $H=3$, and $\eta=0$.

\paragraph{Reporting protocol.}
For trained policies, Tables~\ref{tab:main_results},
\ref{tab:teacher_generalization}, and~\ref{tab:ablation} report
mean $\pm$ sample standard deviation over three seeds, while the
sensitivity figures show the reported success rates. All metrics use a
0--100 scale with higher values indicating better performance, and
success-rate differences are expressed in percentage points. Comparisons
refer to the reported means, without inferring statistical significance
from the three-seed summaries alone.

\subsection{Overall Task Performance}
\label{sec:main_results}

With a Qwen3-32B teacher, Frontier-OPD has the highest reported mean among
trained students on every reported metric at both student scales
(Table~\ref{tab:main_results}). This pattern is consistent with the
supervision-acquisition procedure being useful across the evaluated domains,
although the three-seed summaries do not establish statistical significance.
The larger gaps relative to the compared baselines on ALFWorld and
ScienceWorld for the smaller student are consistent with targeted guidance
being useful when autonomous execution remains difficult, while the
persistent but varying gains for the larger student are consistent with
dependence on the environment and teacher--student configuration.

The comparison with FutureBridge-OPD is more modest on WebShop, where the
small differences in task score and success rate are consistent with gains
in both partial progress and task completion but warrant a more cautious
interpretation than the ScienceWorld results. Some trained students also
outperform the zero-shot teacher, consistent with the value of
environment-specific adaptation; because the teacher is unadapted in this
comparison, however, these results do not establish superiority under
equal adaptation budgets.

\begin{table}[htbp]
\centering
\small
\setlength{\tabcolsep}{4pt}
\renewcommand{\arraystretch}{1.12}
\begin{tabular}{@{}p{0.22\linewidth}*{6}{>{\centering\arraybackslash}p{\dimexpr0.13\linewidth-2\tabcolsep\relax}}@{}}
\toprule
& \multicolumn{3}{c}{\textbf{Qwen3-1.7B student}} & \multicolumn{3}{c}{\textbf{Qwen3-4B student}} \\
\cmidrule(lr){2-4}\cmidrule(l){5-7}
& \textbf{ALFWorld} & \multicolumn{2}{c}{\textbf{WebShop}} & \textbf{ALFWorld} & \multicolumn{2}{c}{\textbf{WebShop}} \\
\cmidrule(lr){2-2}\cmidrule(lr){3-4}\cmidrule(lr){5-5}\cmidrule(l){6-7}
\textbf{Method} & SR $\uparrow$ & Score $\uparrow$ & SR $\uparrow$ & SR $\uparrow$ & Score $\uparrow$ & SR $\uparrow$ \\
\midrule
Student (zero-shot) & 8.2 & 40.5 & 31.3 & 21.1 & 33.0 & 29.3 \\
Teacher (zero-shot) & 82.1 & 64.0 & 62.3 & 82.1 & 64.0 & 62.3 \\
OPD & $55.2 \pm 4.2$ & $60.5 \pm 2.7$ & $58.3 \pm 2.1$ & $67.9 \pm 3.9$ & $64.1 \pm 2.8$ & $64.3 \pm 2.1$ \\
TCOD-F2B & $57.7 \pm 2.2$ & $60.1 \pm 2.4$ & $59.3 \pm 2.1$ & $66.9 \pm 0.9$ & $63.8 \pm 2.8$ & $64.3 \pm 2.1$ \\
TCOD-B2F & $56.5 \pm 2.8$ & $58.8 \pm 2.0$ & $57.7 \pm 1.5$ & $69.4 \pm 1.5$ & $62.9 \pm 1.7$ & $61.3 \pm 1.5$ \\
Guided-OPD & $60.2 \pm 3.5$ & $60.9 \pm 4.1$ & $57.0 \pm 4.4$ & $72.1 \pm 3.4$ & $63.7 \pm 4.2$ & $61.7 \pm 5.0$ \\
FutureBridge-OPD & $61.4 \pm 1.7$ & $60.2 \pm 1.1$ & $58.7 \pm 1.5$ & $70.6 \pm 1.9$ & $63.2 \pm 0.8$ & $62.7 \pm 1.5$ \\
\addlinespace[1pt]
\rowcolor{FrontierBlue}
\textbf{Frontier-OPD (ours)} & $\mathbf{65.9 \pm 3.0}$ & $\mathbf{63.4 \pm 1.5}$ & $\mathbf{61.7 \pm 1.2}$ & $\mathbf{74.6 \pm 2.6}$ & $\mathbf{66.7 \pm 1.5}$ & $\mathbf{65.3 \pm 1.2}$ \\
\bottomrule
\end{tabular}
\caption{Teacher-configuration comparison with Qwen3-8B-RL
teacher and Qwen3-1.7B or Qwen3-4B students. Trained-policy results are
mean $\pm$ sample standard deviation over three seeds. Boldface marks
the highest mean among trained students within each configuration;
blue shading highlights Frontier-OPD.}
\label{tab:teacher_generalization}
\end{table}

\subsection{Teacher Configuration}
\label{sec:teacher_configuration}

To examine whether the results depend on a particular teacher, we repeat
the comparison on ALFWorld and WebShop using Qwen3-8B-RL with both student
scales, following the setting of
\citet{chen2026futurebridge}. Methods within each configuration share the
teacher checkpoint, student initialization, and evaluation protocol,
although the change in both teacher size and training history relative
to Table~\ref{tab:main_results} means that this experiment assesses transfer
across configurations rather than isolating the effect of RL training.

Frontier-OPD retains the highest reported mean among trained students across
all reported metrics (Table~\ref{tab:teacher_generalization}), despite changes
in the relative ordering of the baselines with student size and environment.
Together with the Qwen3-32B results, this consistency is compatible with
the usefulness of selective, continuation-validated guidance across the
evaluated teacher--student pairings, but does not establish performance
outside these configurations.

\subsection{Core Component Ablations}
\label{sec:ablations_efficiency}

We examine the contributions of candidate selection, bridge supervision,
and gain filtering on ALFWorld and WebShop with a Qwen3-8B-RL teacher and
Qwen3-4B or Qwen3-1.7B students (Table~\ref{tab:ablation}). All variants share the ordinary
rollout count, student initialization, reference-prefix curriculum, and
update budget, with $M=4$, $B=4$, $K=1$, $H=3$, and $\eta=0$ defining the full
configuration for both student scales.

\begin{table}[htbp]
\centering
\small
\setlength{\tabcolsep}{3pt}
\renewcommand{\arraystretch}{1.12}
\begin{tabular}{@{}p{0.22\linewidth}*{6}{>{\centering\arraybackslash}p{\dimexpr0.13\linewidth-2\tabcolsep\relax}}@{}}
\toprule
& \multicolumn{3}{c}{\textbf{Qwen3-1.7B student}}
& \multicolumn{3}{c}{\textbf{Qwen3-4B student}} \\
\cmidrule(lr){2-4}\cmidrule(l){5-7}
\textbf{Variant}
& \textbf{ALFWorld} & \multicolumn{2}{c}{\textbf{WebShop}}
& \textbf{ALFWorld} & \multicolumn{2}{c}{\textbf{WebShop}} \\
\cmidrule(lr){2-2}\cmidrule(lr){3-4}\cmidrule(lr){5-5}\cmidrule(l){6-7}
& \textbf{SR $\uparrow$} & \textbf{Score $\uparrow$} & \textbf{SR $\uparrow$}
& \textbf{SR $\uparrow$} & \textbf{Score $\uparrow$} & \textbf{SR $\uparrow$} \\
\midrule
Student (zero-shot) & 8.2 & 40.5 & 31.3 & 21.1 & 33.0 & 29.3 \\
Teacher (zero-shot) & 82.1 & 64.0 & 62.3 & 82.1 & 64.0 & 62.3 \\
w/o bridge data & $62.2 \pm 3.0$ & $58.4 \pm 1.5$ & $58.4 \pm 1.6$
& $70.9 \pm 1.3$ & $61.7 \pm 1.2$ & $62.0 \pm 1.6$ \\
Random selection & $58.7 \pm 3.5$ & $56.0 \pm 1.8$ & $56.7 \pm 1.5$
& $67.4 \pm 2.3$ & $59.3 \pm 1.7$ & $60.3 \pm 1.4$ \\
w/o gain filtering & $63.7 \pm 3.2$ & $60.5 \pm 1.2$ & $59.7 \pm 1.8$
& $72.4 \pm 3.4$ & $63.8 \pm 0.8$ & $63.3 \pm 2.1$ \\
\addlinespace[1pt]
\rowcolor{FrontierBlue}
\textbf{Frontier-OPD (full)} & $\mathbf{65.9 \pm 3.0}$ & $\mathbf{63.4 \pm 1.5}$ & $\mathbf{61.7 \pm 1.2}$
& $\mathbf{74.6 \pm 2.6}$ & $\mathbf{66.7 \pm 1.5}$ & $\mathbf{65.3 \pm 1.2}$ \\
\bottomrule
\end{tabular}
\caption{Core component ablations with a Qwen3-8B-RL teacher and
Qwen3-1.7B or Qwen3-4B students on ALFWorld and WebShop. The full
configuration uses $M=4$, $B=4$, $K=1$, $H=3$, and $\eta=0$. Trained-policy
results are mean $\pm$ sample standard deviation over three seeds; higher
is better.}
\label{tab:ablation}
\end{table}

The \emph{w/o bridge data} variant retains ordinary student experience but
excludes accepted teacher responses from the training pool, testing whether
the additional supervision contributes beyond the base rollout data.
\emph{Random selection} replaces trajectory-balanced node selection with
uniform sampling without replacement from pooled eligible occurrences
under the same candidate budget, whereas \emph{w/o gain filtering} retains
every branch satisfying the shared replay, action-validity, and
continuation-completeness requirements.

The full method achieves the highest reported mean across all reported
metrics for both student scales. Random selection produces the largest
degradation on both benchmarks, suggesting that the allocation of the
intervention budget matters. Removing bridge data also weakens performance,
while accepting all valid bridges remains inferior to the full method despite
outperforming random selection; these patterns are consistent with
complementary effects of acquiring targeted teacher responses and filtering
the resulting guidance before distillation. The comparisons evaluate the complete selection and filtering
procedures: random selection changes both ranking and the allocation unit
(occurrences versus distinct nodes), and the filtering ablation removes
the joint-window acceptance rule, so neither isolates graph structure or
individual terms of the agreement criterion.

\subsection{Sensitivity to Intervention and Trajectory Budgets}
\label{sec:sensitivity}

\begin{figure}[!htb]
\centering
\begin{minipage}[t]{0.49\linewidth}
\centering
\includegraphics[width=\linewidth]{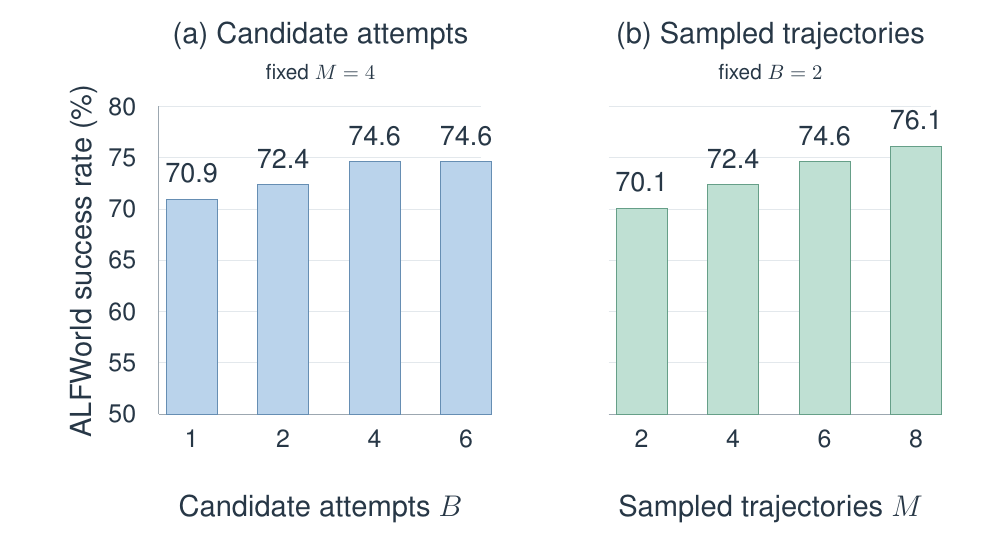}
\captionof{figure}{Sensitivity analysis on ALFWorld with a Qwen3-8B-RL
teacher and Qwen3-4B student. (a) Candidate-attempt budget $B$ with
$M=4$. (b) Sampled trajectory count $M$ with $B=2$. Both studies use
$K=1$, $H=3$, and $\eta=0$. Bars show success rates; the shared vertical
axis spans 50--80\%.}
\label{fig:sensitivity}
\end{minipage}\hfill
\begin{minipage}[t]{0.49\linewidth}
\centering
\includegraphics[width=\linewidth]{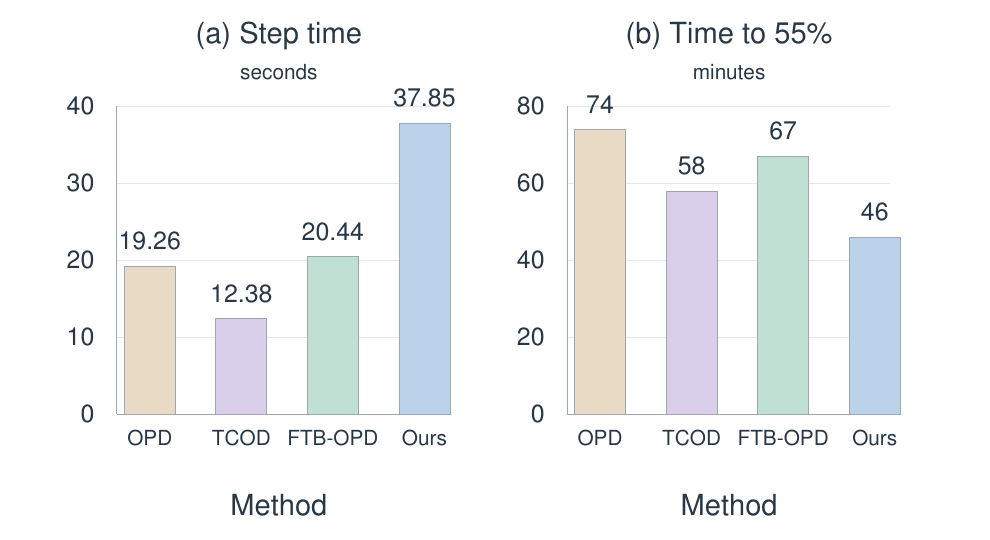}
\captionof{figure}{Training efficiency on ALFWorld with a Qwen3-8B-RL
teacher and Qwen3-4B student on four NVIDIA B300 GPUs. (a) Average
wall-clock time per training step. (b) Elapsed time to reach a 55\% solve
rate. Lower is better for both metrics. FTB-OPD denotes FutureBridge-OPD;
the figure labels Frontier-OPD as ``Ours''.}
\label{fig:efficiency}
\end{minipage}
\end{figure}

Both sweeps use Qwen3-8B-RL $\rightarrow$ Qwen3-4B on ALFWorld with
$K=1$, $H=3$, and $\eta=0$, keeping initialization, training tasks, and
optimization settings fixed within each sweep. When the candidate-attempt
budget increases at a fixed trajectory count
(Figure~\ref{fig:sensitivity}(a)), performance initially increases and then
enters a flatter region. This pattern is consistent with diminishing returns
to additional intervention attempts over the tested range, without
establishing that all useful learning opportunities have been exhausted.

Increasing the sampled trajectory count at a fixed intervention budget
instead produces an upward trend over the evaluated range
(Figure~\ref{fig:sensitivity}(b)). Since additional trajectories enlarge
both the candidate pool and the ordinary distillation dataset, this trend
cannot be attributed to more accurate frontier estimation alone; it
nevertheless suggests that expanding student experience can remain useful
after additional intervention attempts yield limited gains. These sweeps
characterize the tested budget range rather than a universal optimum.

\subsection{Training Efficiency}
\label{sec:supervision_efficiency}

We measure training efficiency on four NVIDIA B300 GPUs using a
Qwen3-8B-RL teacher and Qwen3-4B student on ALFWorld, distinguishing the
average cost of a training step from the elapsed time required to reach
a common 55\% solve rate (Figure~\ref{fig:efficiency}). Although
Frontier-OPD incurs the highest per-step cost because of its additional
collection and validation work, it reaches this target in the least elapsed
time in this measurement. The result is consistent with a time-to-target
advantage in this setting, while leaving open how the tradeoff changes
at other success thresholds or hardware scales.

\section{Conclusion}
\label{sec:conclusion}

We presented Frontier-OPD, a framework that organizes supervision
acquisition around the student's learning frontier by jointly considering
where guidance is needed and whether the current policy can build on it.
Trajectory-balanced aggregation identifies candidate decision situations
from shared-state experience, while matched-window validation screens
short teacher bridges before their responses enter on-policy distillation.
By renewing this selection process with fresh student rollouts, the method
adapts supervision to the evolving policy and yields higher reported mean
autonomous performance across the evaluated comparisons on three interactive
benchmarks. These results motivate
further investigation of learning-frontier dynamics, particularly through
direct measurements of whether situations that initially require teacher
guidance become independently solvable after training.

\section*{Limitations}
Our learning-frontier estimate relies on behavioral proxies whose relation
to acquired capability remains imperfect: high disagreement need not
indicate a missing skill, and non-decreasing agreement over a joint
bridge-and-continuation window guarantees neither better continuation
alone nor subsequent mastery. Although the experiments measure autonomous
task performance, they do not directly trace skill acquisition at selected
nodes or establish monotonic expansion of the learning frontier.

The estimate is also conditioned on sampled experience and an
observation-based state abstraction, which can merge histories with
different learning needs and admits nodes with only one eligible visit.
Shared state identity influences candidate ranking, but graph adjacency
does not enter the score; moreover, because the random-selection ablation
changes both ranking and the allocation unit, it cannot isolate the
benefit of state aggregation. Practical applicability depends on replayable
environments and access to teacher token probabilities, while the measured
timing advantage is limited to one target success rate on ALFWorld and
may vary with the performance threshold and hardware scale. The reported
trained-policy comparisons use three seeds and sample standard deviations;
they do not establish statistical significance or robustness to additional
seeds. The Qwen3-32B and Qwen3-8B-RL experiments also change teacher size
and training history, so their results should not be read as an isolated
teacher-scaling or RL-training comparison. Finally, all evidence comes from
three benchmark environments and does not establish generalization to
unrestricted real-world agents.

\bibliographystyle{assets/plainnat}
\bibliography{frontier_opd}

\clearpage
\beginappendix
\section{Evaluation Protocol}
\label{sec:experiment_appendix}
\label{sec:evaluation_protocol}

Students are evaluated on held-out task instances from the environment's
initial state, without teacher intervention or reference-prefix replay,
using the same task identifiers, action parser, prompts, history
construction, and decoding configuration within each controlled comparison.
Checkpoint selection is fixed before test evaluation, and each run retains
its configuration, seed, checkpoint identifier, and per-task records, with
task errors distinguished from unsuccessful but valid episodes.

\paragraph{Teacher configurations.}
\label{sec:generalization}
The main comparison in Table~\ref{tab:main_results} uses a Qwen3-32B teacher,
whereas Table~\ref{tab:teacher_generalization} uses Qwen3-8B-RL, both with
Qwen3-1.7B and Qwen3-4B students. The core ablations,
budget-sensitivity studies, and training-efficiency comparison use
Qwen3-8B-RL $\rightarrow$ Qwen3-4B, with a shared teacher checkpoint within
each controlled comparison and checkpoint identifiers and RL-training
provenance retained in the experimental records.

The concrete benchmark configurations used by the workflows are summarized
in Table~\ref{tab:protocol}. Training uses the train split and has no
separate validation split; test evaluation is performed sequentially on the
held-out split. The reported three-seed aggregates use the same configuration
for every method within a comparison, while the rollout configuration uses
seed 42. Evaluation starts from the environment's initial state and uses
temperature $0.4$; training rollouts and teacher rescoring use temperature
$1.0$. Qwen3 thinking mode is disabled throughout. The linear backward-to-
forward checkpoint schedule uses five training steps per prefix shift, and
the final reported checkpoint is step 200.

\begin{table}[!htbp]
\centering
\caption{Benchmark-specific evaluation and training settings used by the
Frontier-OPD workflows.}
\label{tab:protocol}
\scriptsize
\setlength{\tabcolsep}{2.5pt}
\renewcommand{\arraystretch}{1.04}
\begin{tabular*}{\linewidth}{@{\extracolsep{\fill}}lccc@{}}
\toprule
Setting & ALFWorld & WebShop & ScienceWorld \\
\midrule
Environment turn limit & 30 & 15 & 30 \\
Maximum prompt tokens & 10,240 & 10,240 & 10,240 \\
Maximum response tokens & 512 & 512 & 512 \\
Training temperature & 1.0 & 1.0 & 1.0 \\
Evaluation temperature & 0.4 & 0.4 & 0.4 \\
Evaluation task-set budget & 134 steps & 100 steps & 1,308 steps \\
Rollout model seed & 42 & 42 & 42 \\
Auxiliary model seed & 42 & 42 & 42 \\
Prefix curriculum & linear B2F & linear B2F & linear B2F \\
Prefix shift interval & 5 steps & 5 steps & 5 steps \\
Optimizer & AdamW & AdamW & AdamW \\
Learning rate & $10^{-6}$ & $10^{-6}$ & $10^{-6}$ \\
Training steps & 200 & 200 & 200 \\
Collection batch size & 4 & 16 & 4 \\
Training batch size & 32 & 64 & 32 \\
\bottomrule
\end{tabular*}
\end{table}

\section{Frontier-OPD Algorithm}
\label{sec:appendix_algorithm}

Algorithm~\ref{alg:frontier_opd} summarizes supervision acquisition for a
batch of tasks using the notation of Section~\ref{sec:method}. All sampling
and scoring within an iteration use the same student policy
$\pi_{\theta_k}$, with graphs and candidate pools constructed separately
for each task before accepted experiences are aggregated for the update.

\newcounter{frontieralgorithm}
\begin{table}[htbp]
\centering
\refstepcounter{frontieralgorithm}\label{alg:frontier_opd}
\begin{minipage}{\linewidth}
\hrule\vspace{5pt}
\textbf{Algorithm \thefrontieralgorithm: Frontier-OPD}
\par\vspace{4pt}\hrule\vspace{5pt}
\textbf{Input:} student $\pi_{\theta_0}$, teacher $\pi_{\mathrm T}$, training
tasks, budgets $M,B,K,H$, acceptance margin $\eta$.\par
\textbf{Output:} distilled student $\pi_{\theta}$.\par
\begin{tabular}{@{}r@{\quad}p{\dimexpr\linewidth-2em\relax}@{}}
1 & \textbf{for} collection iteration $k=0,1,\ldots$ \textbf{do} \\
2 & \quad Initialize $\mathcal D_k^{\mathrm{stu}}\gets\emptyset$ and
    $\mathcal D_k^{\mathrm{br}}\gets\emptyset$. \\
3 & \quad \textbf{for} task $x$ in the current batch \textbf{do} \\
4 & \qquad Sample $M$ student trajectories from the prescribed starting histories. \\
5 & \qquad Store ordinary student responses in $\mathcal D_k^{\mathrm{stu}}$;
    retain contexts, actions, and both models' token log-probabilities. \\
6 & \qquad Construct $\mathcal G_k(x)$ from shared state keys and observed transitions. \\
7 & \qquad Group eligible visits by node and trajectory; compute $s_k(v)$
    using trajectory-balanced weights in \eqref{eq:node_score}. \\
8 & \qquad Select top-$B$ distinct nodes $\mathcal F_k$ by $s_k(v)$ after
    the node threshold; set $\mathcal C_k=\{c_k^*(v):v\in\mathcal F_k\}$. \\
9 & \qquad \textbf{for} representative $c\in\mathcal C_k$ \textbf{do} \\
10 & \qquad\quad Restore the original task and replay the prefix preceding $c$;
     skip the candidate if replay fails. \\
11 & \qquad\quad Execute the teacher from the saved context for at most $K$ turns,
     stopping at graph exit or environment termination. \\
12 & \qquad\quad Let $L_c$ be the realized bridge length; reject invalid bridges
     or terminal teacher handoffs. \\
13 & \qquad\quad Resume the student for $H$ turns; reject candidates without a
     complete valid continuation and a matched reference window. \\
14 & \qquad\quad Form $W_c^{\mathrm{new}}$ from the bridge and continuation;
     take $W_c^{\mathrm{old}}$ from the original $L_c+H$ response turns. \\
15 & \qquad\quad Compute $\Delta_k(c)\gets A_k(W_c^{\mathrm{new}})-A_k(W_c^{\mathrm{old}})$. \\
16 & \qquad\quad \textbf{if} $\Delta_k(c)\geq\eta$ \textbf{then} add all teacher
     bridge responses to $\mathcal D_k^{\mathrm{br}}$. \\
17 & \qquad \textbf{end for} \\
18 & \quad \textbf{end for} \\
19 & \quad $\theta_{k+1}\gets\operatorname{OPDUpdate}
     (\theta_k;\mathcal D_k^{\mathrm{stu}}\uplus\mathcal D_k^{\mathrm{br}})$. \\
20 & \textbf{end for}
\end{tabular}
\par\vspace{5pt}\hrule
\end{minipage}
\end{table}

The candidate budget is allocated to distinct graph nodes, each supplying
exactly one eligible replay representative, although several nodes may
select representatives from the same trajectory. Every intervention is
replayed independently from its original prefix, so an accepted bridge
does not alter another candidate's reference; newly generated student
continuations are used exclusively for validation and are excluded from
the training pool.

\section{Implementation Details}
\label{sec:appendix_implementation}

\paragraph{State abstraction and decision identity.}
State keys combine normalized observations with available-action
information, using deterministic normalization and ordering to avoid
creating distinct keys from incidental formatting differences. Action
information consists of admissible commands in ALFWorld, search
availability and clickable elements in WebShop, and valid action--object
combinations in ScienceWorld. Although the key determines node-score
aggregation and graph membership, each occurrence retains its original
context, response, and replay prefix, allowing disagreements from distinct
histories to contribute to a shared-node estimate without replacing the
model input. This observational abstraction need not distinguish every
latent environment state.

\paragraph{Candidate eligibility and ranking.}
An eligible occurrence must have valid actions through the candidate, a
replayable prefix, a state key, aligned nonempty token-score arrays, finite
disagreement, and enough original responses for the $K+H$ reference horizon.
Equation~\eqref{eq:node_score} is implemented by first averaging eligible
disagreements within each trajectory at a node and then averaging over
contributing trajectories, so neither visit count nor token count determines
a trajectory's total weight. For example, three visits with score 10 from
one trajectory and one visit with score 0 from another yield a node score
of 5 rather than 7.5. Negative scores remain valid, and the optional
minimum-disagreement threshold is applied only after aggregation, before
the selector takes at most $B$ distinct nodes and assigns each its
highest-disagreement eligible replay representative. There is no
per-trajectory candidate quota; occurrence ties follow trajectory/turn
order, while node-score ties follow the first eligible encounter.

\paragraph{Replay and handoff.}
Before each intervention, the environment is restored and the recorded
student actions are replayed, preceded by the reference prefix when a
curriculum is active, after which teacher generation uses the saved
decision context. A bridge lasts at most $K$ response turns and may stop
earlier at graph exit, but terminal handoffs are excluded because
validation requires subsequent student behavior. At the default $K=1$,
both graph-internal and graph-external handoffs remain eligible, meaning
that shared-node aggregation still affects intervention selection even
though graph novelty is not an acceptance condition.

\paragraph{Agreement computation.}
Both models score the same tokens under each response's own context,
including teacher-generated bridge responses, with agreement computed as
the negative mean absolute log-probability gap within a response and then
averaged uniformly across turns. This normalization prevents longer
responses from receiving greater weight solely because of their length,
while matching $L_c+H$ turns on each branch uses the realized bridge length
$L_c$ rather than its upper bound $K$. Reference scores are reused from
the collected trajectory, and acceptance requires all validity conditions
in \eqref{eq:accepted_bridges} together with a gain of at least $\eta$,
including equality when $\eta=0$.

\paragraph{Training data and renewal.}
Ordinary student responses are retained regardless of bridge acceptance,
and all responses in accepted teacher bridges enter the same distillation
pool, with repeated occurrences preserved rather than collapsed by graph
identity. Excluding the validation continuations separates the behavioral
evidence used to select supervision from the additional supervision itself.
Subsequent rollouts reconstruct the graph and candidate scores, adapting
the supervision distribution to the updated policy without presuming that
every accepted candidate has become mastered.

\paragraph{Default settings.}
The default configuration uses $M=4$, $B=4$ distinct nodes, $K=1$, $H=3$,
and $\eta=0$, with
\texttt{frontier\_node\_weighting=}\allowbreak\texttt{trajectory\_balanced}
and no per-trajectory candidate quota.

\paragraph{Collection cost.}
For $N$ collected occurrences and $U$ eligible distinct nodes, graph assembly
and trajectory-balanced aggregation require $O(N)$ time apart from state-key
normalization, followed by $O(U\log U)$ sorting to select the first $B$
nodes and $O(N+U)$ storage in the selector. Beyond ordinary rollouts and
prefix replay, validation requires at most $B(K+H)$ additional generation
turns per task and the corresponding probability evaluations, whereas at
most $BK$ teacher responses enter training. Since rejected candidates
consume collection resources without contributing bridge data, collection
cost and accepted supervision volume must be considered separately.

\section{Extended Related Work}
\label{sec:appendix_related_work}

\paragraph{Distillation and the supervision distribution.}
Knowledge distillation transfers predictive information through teacher
outputs~\citep{hinton2015distilling}, while sequence-level distillation
extends the target to teacher-generated sequences for autoregressive
students~\citep{kim2016sequence}. These formulations distinguish the
supervision target from the distribution of sequences on which it is
applied, a distinction central to on-policy distillation's use of
student-generated responses~\citep{agarwal2024opd} and MiniLLM's study of
on-policy reverse-KL optimization~\citep{gu2024minillm}. Frontier-OPD
addresses the acquisition of additional supervision within interactive
trajectories, retaining ordinary student experience while selectively
adding short teacher responses.

\paragraph{Interactive imitation and curricula.}
Because learner actions determine the states encountered during execution,
dataset aggregation collects expert supervision under learner-induced
state distributions~\citep{ross2011dagger}, while scheduled sampling
addresses a related training--inference mismatch by gradually replacing
ground-truth prefix tokens with model-generated ones
\citep{bengio2015scheduled}. Agentic OPD extends the setting to
language-conditioned environment histories, with TCOD adjusting student
execution depth~\citep{wang2026tcod} and Guided-OPD scheduling teacher
participation at individual turns~\citep{li2026guidedopd}. Frontier-OPD's
aggregation of decision-level disagreement and history-specific validation
operate at the level of supervision allocation, allowing both to coexist
with curricula that determine the starting histories of student rollouts.

\paragraph{Feedback in multi-turn agents.}
Environment observations and linguistic feedback support sequential
behavior through approaches such as ReAct's interleaving of reasoning and
actions~\citep{yao2023react} and Reflexion's use of episodic memory to
improve later trials without parameter updates~\citep{shinn2023reflexion}.
Within distillation, FutureBridge-OPD evaluates guidance through its induced
student continuation~\citep{chen2026futurebridge}, providing the rationale
for assessing interventions beyond the teacher's immediate response.
Frontier-OPD combines this perspective with trajectory-balanced candidate
ranking: shared observations organize evidence of learning needs, and
matched-window agreement determines which teacher responses contribute
to the subsequent parameter update.

\paragraph{Interactive evaluation.}
AgentBench's evaluation across diverse interactive environments
\citep{liu2024agentbench} and WebArena's assessment of functional task
completion on reproducible websites~\citep{zhou2024webarena} emphasize
capabilities that extend beyond isolated response quality. ALFWorld,
WebShop, and ScienceWorld provide complementary settings for evaluating
such grounded sequential behavior
\citep{shridhar2020alfworld,yao2022webshop,wang2022scienceworld}.
Our evaluation measures autonomous execution after training, keeping
teacher interventions confined to supervision acquisition and distinguishing
task-level performance from the agreement scores used to select that
supervision.

\section{Benchmark Descriptions}
\label{sec:appendix_benchmarks}

\paragraph{ALFWorld.}
ALFWorld aligns text-based interaction with embodied household tasks
\citep{shridhar2020alfworld}; we use its text interface, in which agents
receive instructions and observations and act through admissible commands.
Coordinating navigation, object interaction, and task-specific conditions
makes this setting suitable for examining whether localized guidance
improves decisions under changing object and location contexts, with
success rate measuring task completion rather than reproduction of a
particular demonstration.

\paragraph{WebShop.}
WebShop combines real-world product information with natural-language
shopping requests~\citep{yao2022webshop}, requiring agents to search,
navigate product pages, select options, and complete purchases that satisfy
the requested attributes. Task score captures graded fulfillment and
success rate captures completion under the evaluation criterion, while
page observations and available clickable elements provide the
observational identity used to organize revisited states.

\paragraph{ScienceWorld.}
ScienceWorld evaluates scientific reasoning through interaction in a
text-based simulated world~\citep{wang2022scienceworld}, where agents must
translate goals into executable sequences involving objects, tools, and
environmental changes. This setting tests whether local guidance can be
followed by further student-controlled procedural reasoning, with task
score capturing partial progress and success rate measuring completion
of the objective.

\paragraph{Benchmark selection.}
The suite spans household manipulation, web information seeking, and
scientific procedures, providing domains that share policy-dependent
observations but differ in action interfaces, task structure, and feedback.
The main comparison covers all three benchmarks, while the RL-teacher
comparison and core ablations use ALFWorld and WebShop, and the sensitivity
and efficiency studies use ALFWorld. This scope permits evaluation of a
common supervision-acquisition method across distinct environments,
without establishing generalization to unrestricted real-world settings.

\section{Reproducibility Statement}
\label{sec:reproducibility}
Section~\ref{sec:method} specifies the selection, replay, validation, and
update rules needed to reproduce the method, while
Section~\ref{sec:experimental_setup} defines the benchmarks, comparisons,
and default intervention settings. The ablation and sensitivity sections
identify the controlled changes, with evaluation conditions and
teacher--student configurations detailed in
Appendix~\ref{sec:evaluation_protocol}.

\end{document}